\pdfoutput=1
\documentclass[11pt]{article}

\usepackage{ACL2023}

\usepackage{times}
\usepackage{latexsym}
\usepackage{booktabs}
\usepackage{tabularx}
\usepackage{footmisc}

\usepackage[T1]{fontenc}
\usepackage[utf8]{inputenc}

\usepackage{microtype}

\usepackage{inconsolata}

\title{Political claim identification and categorization in a multilingual setting: First experiments}

\author{Urs Zaberer \and Sebastian Padó \and Gabriella Lapesa \\
            Institut für Maschinelle Sprachverarbeitung \\
            University of Stuttgart, Germany  \\
            \texttt{\{urs.zaberer,sebastian.pado,gabriella.lapesa\}@ims.uni-stuttgart.de}}

\begin{document}
\maketitle
\begin{abstract}
% Problem setting.
The identification and classification of political claims is an important
step in the analysis of political newspaper reports; however,
resources for this task are few and far between.
This paper explores different strategies for the cross-lingual
projection of political claims analysis. We conduct experiments on a
German dataset, DebateNet2.0, covering the policy debate sparked by
the 2015 refugee crisis. Our evaluation involves two tasks (claim
identification and categorization), three languages (German, English,
and French) and two methods (machine translation -- the best method in
our experiments -- and multilingual embeddings).
\end{abstract}

\section{Introduction}

% Why we need political claims

The identification of political claims in news is a core step in
the % text-as-data
analysis of policy debates. \textit{Discourse
networks}, whose nodes correspond to claims and the actors who advance
them, provide a rich source of information on phenomena such as
formation of coalitions (who agrees with whom), shift in salience due
to external events (e.g., migration waves making the issues of
refugee accommodation more central in a debate), emergence of
leadership, and polarization of a discourse
\citep{leifeldPoliticalDiscourseNetworks2012,koopmansPoliticalClaimsAnalysis1999,hajerDiscourseCoalitionsInstitutionalization1993}.

Political claims are defined as demands, proposals or
criticism that are \textit{supported} or \textit{opposed} by an
\textit{actor} (a person or a group of persons). Political claims
generally form a call to action: they refer to something that should (or
should not) be done in a policy domain (e.g., assigning empty
flats to refugees). Thus, political claims are related to, but add a
new perspective on, the Argument Mining question of what
claims are, and what are the best strategies for modeling them across
domains
\citep{daxenberger-etal-2017-essence,schaefer-etal-2022-selecting}.

The potential and challenges of the NLP support to political claim
analysis have been thoroughly explored in the recent years in a
monolingual setting
\citep{chen-etal-2020-detecting,dayanik-etal-2022-improving}; however,
there are very few resources available in multilingual or
crosslingual settings. Thus, there is little work on the comparison of
policy debates in different countries, either completely automatic, or
semi-automatic (supporting the inductive development of annotation
guidelines in a new language).

This paper reports on cross-lingual pilot experiments on two tasks
(claim identification and categorization), comparing two well known
approaches to cross-lingual transfer in NLP in general, and argument
mining in particular: machine translation and multilingual embeddings
\citep{eger-etal-2018-cross,toledo-ronen-etal-2020-multilingual}. We
first work with a reference dataset for the German migration policy
debate \citep{blokker:_between}, and on its projection to English and
French, before moving on to a newly annotated English test set on the
same topic. Machine Translation turns out to be the best cross-lingual
projection strategy.

\section{Experimental Setting}

\subsection{Tasks} \label{tasks}

This work focusses on two constituent tasks of political claim
analysis \cite{pado-etal-2019-sides}. Our first task is \textbf{claim
identification}, performed as a binary classification task at the
sentence level. Our second task is \textbf{claim
categorization}, phrased as a multi-label classification task at the
sentence level.\footnote{For our evaluation in the claim categorization task, we consider all claims in the manually annotated gold standard.} 

\subsection{Data}\label{data}

We carry out two experiments. In the first one, we use a German
corpus, DebateNet, which we automatically translate into English and
French: this represents a cross-lingual transfer within the same media
outlet. In the second experiment, we transfer our DebateNet models to
an original English dataset based on the \textit{Guardian} newspaper.

\paragraph{DebateNet 2.0.} \citet{blokker:_between} is a dataset\footnote{http://hdl.handle.net/11022/ 1007-0000-0007-DB07-B} targeting the German
public debate on migration policies in the context of the 2015
so-called 'refugee crisis'. It is based on 700 articles from the
German quality newspaper \textit{die Tageszeitung (taz)} with a total
of 16402 sentences.

% refugee crisis in 2015 in Germany. 
% \textit{DebateNet2.0} is publicly available as a CLARIN
% resource
% and contains annotations on articles covering the entire year 2015. Of the
% 38k articles from 2015, 700 have been selected as related to the
% migration topic, combining a keyword-based and automatic selection
% approach.

%DebateNet annotates both political claims and the correspoding
%actors.
Political claims are annotated as textual spans, and each claim span
is associated with at least one of 110 categories drawn from a
theory-based codebook (annotation guidelines). Around 15\% of
sentences are annotated to contain a claim span. In total, the dataset
contains 3442 claim spans corresponding to 4417 claim labels (i.e.,
each claim span is associated with an average of 1.3 claim
categories). Annotations are first proposed by pairs of students of
political science, with an inter-coder reliability is $\kappa=0.59$
\citep{pado-etal-2019-sides}, and then accepted, rejected or merged by
domain experts. We randomly split DebateNet into a training,
development, and test set with a ratio of 80:10:10.

Crucially for our experiments, the 110 fine-grained categories are
organized into 8 top-level categories which encode general domains
of the migration policy field. In the claim categorization experiments in this paper we focus on the 8 top-level categories. Table \ref{tabelle_classes} in the
Appendix shows them with the percentage of claims annotated for each
category and illustrative examples.

% (see \citet{blokker:_between} for details). 

% The core unit of annotation are \textit{political claims}, defined as
% request of specific actions to be taken in the policy field (e.g.,
% assigning empty flats to refugees; establishing quotas for refugees to
% be admitted in the country; increasing integration offers for
% refugees) and the \textit{political actors} who make them
% (politicians, parties, or less institutional groups, such as
% protesters). As the focus of this paper is on political claims, we
% refer the reader to  for a detailed
% description of the actor annotation (and additional layers) contained
% in \textit{DebateNet2.0}.

\paragraph{Guardian test set}
%was: translated into German, probably a typo
To compare German news translated into English to actual UK news, we
collected an English-language test set of 36 articles from the British
quality newspaper Guardian, extracted from the World News section and
published in 2015. To make our test set as compatible as possible with
\textit{DebateNet2.0}, we look at the five months most represented in
\textit{DebateNet2.0} and within each month sample from articles
written in the seven-day spans with the highest frequency of articles
in \textit{DebateNet2.0}.
% removed bc that's only 4 months September, October, November, January. 
Articles were further filtered by keywords
(\textit{migrant, refugee, asylum, Germany, Syria, Afghanistan} and
their morphological and syntactic variants) and by the mention of the
most salient political actors (politician and parties).

 The Guardian test set was manually annotated by a native speaker, a
 MSc-level student in Computational Linguistics, based on the
 \textit{DebateNet2.0} guidelines. Claims were
 identified and assigned to one of the 8 top-level categories
 described in the previous section. Across the 36 articles
 with 1347 sentences, the test set contains 82 claim spans which
 correspond to 101 claim categories (mean of 1.2 categories per
 span).\footnote{30 claims, albeit identified by our annotator, could
 not be classified in any categories of the codebook.} Refer to Table
 \ref{tabelle_classes} in the Appendix for the distribution of claim
 categories.

% For the claim categorization
%task, we rely on a dataset 82 claim spans, corresponding to 101
%individual claims.
%attributed to 36 distinct actors (31 politicians and 5 parties). 

\subsection{Methods}\label{trainingsetup}

\subsubsection{Projection methods}

With the German DebateNet2.0 as our starting point, and the goal of
testing the feasibility of cross-lingual projection to English and
French (as target languages), we compare the two most established
projection methods
\citep{eger-etal-2018-cross,toledo-ronen-etal-2020-multilingual}:
machine translation (to make the modeling task monolingual) and
multilingual embeddings (to let the model bridge the language gap
implicitly). This yields three experimental conditions:

\paragraph{Translate-train:} We machine-translate the German training data into the
target languages and fine-tune a monolingual target-language model on
it, to be evaluated on the target-language test data.\footnote{We uses the DeepL
translator via its web interface on a free trial of the “advanced”
plan as of August 2022.}

\paragraph{Translate-test:} We machine-translate the test data
into German (as described above) and apply a monolingual German model
fine-tuned on the original German data to it.  For the DebateNet
experiments in Section \ref{experiments-debatenet}, we can only
simulate this setting, as we do not have genuine foreign-language test
data. We simulate it with a back-translation: first, we
machine-translate the German DebateNet test set into the target
language (EN/FR); then we translate the simulated EN/FR test sets back
into German. It is only on the Guardian test set (Section\
\ref{sec:guardian-test-set}) that we can fully evaluate our models in
the translate-test configuration.

\paragraph{Multilingual:} We employ multilingual embeddings,
fine-tune them on the original German data, and apply the resulting
classifier on the target language test data, exploiting the model's
internal alignment of the source and target languages.

\hspace{\baselineskip}

\noindent For both claim identification and classification, we
re-implement standard Transformer-based models from the literature
\citep{dayanik-pado-2020-masking}. We use BERT as well as its German,
French and multilingual versions. Details on the classifier setups
for both tasks follow below.

%For both tasks we employed the \verb|BertForSequenceClassification|
%function from Huggingface. Details on the training and test
%configurations, selected monolingual embeddings, as well as
%hyperparameter settings, are provided below.

\subsubsection{Claim identification}

\paragraph{Translate-train:} For English, we select the uncased model (\verb|bert-base-uncased|) based on its performance on the development set, and we set learning rate to 5e-5 and warm-up steps to 30. The same configuration is used for the German monolingual baseline.
For French, we select the base version of CamemBERT, \verb|camembert-base|, with a learning rate of 4e-5 with 30 warm-up steps.

\paragraph{Translate-test:} we employ a German BERT model, \verb|bert-base-|\verb|german-|\verb|cased|, fine-tuned on the original German dataset. The hyperparameters are the same as for English translate-train.

\paragraph{Multilingual:} Based on performance on the development set, we select the cased variant of the multilingual BERT from the Huggingface transformer library, \verb|bert-base-multilingual-cased|. Training this model requires a lower learning rate of 2.5e-5 and correspondingly more epochs.

\subsubsection{Claim categorization}

\paragraph{Translate-train: } For the English model, we assess
both the cased and uncased versions. Since the uncased one
(\verb|bert-base-uncased|) again performs slightly better, we select
it and use a learning rate of 5e-5. Experiments on the corresponding
development sets establishes 25 warm-up steps as a reasonable choice
for all configurations in Task 2. The French model – the same as for
the claim identification task – requires a learning rate of 4e-5.

\paragraph{Translate-test: } We employ \verb|bert-|\verb|base-| \verb|german-|\verb|cased| with a learning rate of 4e-5.
The same model is also used for the monolingual German baseline model.

\paragraph{Multilingual: } Based on performance on the development set,
we select \verb|bert-base-| \verb|multilingual-uncased| with a low learning rate of 3e-5
and correspondingly more epochs.

\section{Experiment 1: Within-outlet cross-lingual transfer}

\begin{table}[tb!]
	\begin{center}
		\begin{tabular}{lrrll}
			\toprule
			Setup & Train & Test & Id &  Cat \\
                        \cmidrule(r){1-3}
                        \cmidrule(l){4-5}
%                           & & & F1 & F1 \\                        
%			\midrule
   			BL (mono) & de & de & 56.2   & \textbf{70.5}   \\
                        \cmidrule(r){1-3}
                        \cmidrule(l){4-5}
%           \midrule
   			Translate-train & en & en & \textbf{57.3}  & 67.8  	\\
			Translate-train & fr & fr & \textbf{57.4}  & 69.7  	\\
   			%translate-train & bad en & en & 54.8\textdagger\\
                        \cmidrule(r){1-3}
                        \cmidrule(l){4-5}
%          \midrule
			Translate-test & de & de-en & 55.8  & 69.5 	\\
			Translate-test & de & de-fr & 58.3  & 69.8  	\\
                       \cmidrule(r){1-3}
                        \cmidrule(l){4-5}
%		      \midrule
			Multilingual & de & en & 45.8  & 50.3  	\\
			Multilingual & de & fr & 51.1  & 51.0  	\\
                       \cmidrule(r){1-3}
                        \cmidrule(l){4-5}
%     		Multilingual\textdagger\footref{f4} & de & de & 55.6  & 63.3    \\ 
            Multilingual\textdagger & de & de-en & 52.0  & 60.0 	\\  
      		Multilingual\textdagger & en & de & 55.4  & 64.1     \\	 
            \bottomrule
		\end{tabular}
		\caption{DebateNet test set results: F1 scores (positive class for claim identification (ID), macro average for claim categorization (Cat)). BL (mono): monolingual baseline.}
                \label{tabelle_id}
	\end{center}
\end{table}	

\subsection{Claim Identification on DebateNet}\label{experiments-debatenet}

The left-hand side of Table \ref{tabelle_id} shows results for the
first main experiment, comparing the translate-train, translate-test,
and the multilingual embedding approaches to claim identification to a
monolingual baseline.\footnote{Unless indicated by a dagger
\textdagger, reported values for all conditions are the averages of two runs to reduce
variance. \label{f4}} For comparison, we also run the translate-train
and translate-test approaches on the multilingual model
(multilingual:en:de and multilingual:de:de-en). The language labels
de-en and de-fr stand for German data translated into EN or FR and
back-translated into German.

The main contrast of this set of experiments is the one between the
translate-train approach and the multilingual embeddings approach
%(and
%their relation to the monolingual baseline, of course)
with respect to their performance on the target languages (EN/FR). For
both target languages, the translate-train approach outperforms the
monolingual baseline and the multilingual embedding approach. We
ascribe this (small) performance gain to the higher quality of the
embeddings available for the target languages: 
%\footnote{While the
%models used in our experiments all use the same basic architecture,
%model results still depend on the size of the corpus they are trained
%on.
The monolingual English model, \texttt{bert-base}, is trained on a
much larger corpus (English Wikipedia and BookCorpus) than
\texttt{bert-base-german}, which is only trained on the significantly
smaller German Wikipedia. The French model’s training corpus is also
over ten times larger than the German one. This also means the
translation process, albeit not perfect, has not degraded the claim
"signal" in the training data.

This point is also supported by the results for the "simulated"
translate-test approach, which (cf. Section\ \ref{trainingsetup}) can
be considered a test of translation quality. Since the performance is
in line with the monolingual baseline (de-en) or even slightly
superior to it (de-fr)\footnote{The exact reason for the improved
performance in the de-fr setup is to be further investigated. Given that we consider the translate-test setup in DebateNet as a translation quality check, the result is not highlighted in bold even if higher than translate-train.}, the
claim signal is preserved through the back-translation process.
%\marginpar{Seb: why is 58.3 not boldface even though it's the best?}

In contrast, the multilingual embeddings perform poorly, below the
monolingual baseline. The bottom part of Table \ref{tabelle_id} shows
additional experiments we carried out to better understand this
result. We find that a monolingual setup with multilingual embeddings
(DE-DE) still performs below the monolingual baseline, but the
performance gap is narrower than for the cross-lingual setups (DE-EN
and DE-FR). Reverting the direction of the mapping, contrasting the
performance of English-German (55.6) vs. German-English (45.8), again
speaks in favor of the German representations being the weak point --
the training data for the English-German multilingual embeddings setup
is the same as that of the translate-train approach.

\begin{table}[tb]
	\begin{center}
		\begin{tabular}{rrr}
			\toprule
			 & Target: yes & Target: no  \\
			\midrule
			Predicted: yes & 71 & 39\\	
			Predicted: no & 75 & 822\\
			\bottomrule
		\end{tabular}
		\caption{Claim identification (DebateNet) confusion matrix of the best model for English (translate-train)}\label{tabelle_conf}
	\end{center}
\end{table}

The confusion matrix for the best cross-lingual model for English
(translate-train), Table \ref{tabelle_conf}, shows
many fewer false negatives than false positives (i.e., a high
precision). Regarding application to the (semi-)automatic extraction
of discourse networks, this outcome is complementary to the
high-recall approach applied by
\citet{haunss20:_integ_manual_autom_annot_creat} to the German
annotation in DebateNet, but lends itself to high-precision
human-in-the-loop approaches like the one proposed by
\citet{EinDor2019CorpusWA} for argument mining.

\paragraph{Error Analysis.} The misclassified instances provide some more
insight into the model. For instance, we might expect the word
“fordern” (“demand”, “call for”) to frequently appear in claims and
therefore lead the model to make a positive prediction. Indeed, in the
misclassified instances of the German-French translate-test model,
forms of the word “fordern” or “Forderung” are 13 times more likely to
be FP than FN even though there are almost twice as many FNs. We
can therefore conclude that this word influences the model in the
expected way. We bolster these observations with more formal methods:
using saliency-based analysis \citep{DBLP:journals/corr/SimonyanVZ13}
% as implemented in the \texttt{transformers-interpret} package,
we can assign each token a relevance for the model's prediction.  The
results partially confirm this: the token “fordert” gets
scores above 0.9 throughout. However, other forms, like the infinitive, receive lower scores, presumably because the 3rd person singular is more highly associated with concrete claiming situations.

Saliency scores are highly correlated between models and between
languages. E.g., the sentence “Der bayerische Ministerpräsident Horst
Seehofer begrüßte die Pläne” and its corresponding English version
‘Bavaria’s prime minister Horst Seehofer welcomed the plans.’, are
both labeled as claims. In both cases, the highest saliency is
assigned to “Pläne”/”plans”. A systematic comparison of scores among
models is however complicated by the differences in tokenizations among
embedding models.
%Comparing the scores assigned by
%different models  raises problems, though, since there
%is often no one-to-one mapping between tokens.
%For the “Pläne”/”plans”
%pair, for example, the Spearman correlation between models is 0.80,
%but results for other sentences are more mixed with less intuitive
%attention spreads both on sentences classified correctly and
%incorrectly.
Alternatively, we can compare
instances misclassified by different models. Here, we observe large
overlap. On one test run, the multilingual German-French model
misclassified 122 out of 1007 test instances, while the monolingual
English model misclassified 120 instances. These instances have an
overlap of 58\% (random assignment, should result in 12\%
overlap). This suggests that the models struggle with the same
instances. A first qualitative inspection at such "difficult"
instances has ruled out the impact of proper names, length of
sentences as well as the type of involved actors; further analysis in this
direction is required.
%\marginpar{This is nice to a bit inconclusive; can go if we need space}

\subsection{Claim Categorization on DebateNet}

The right-hand side of Table \ref{tabelle_id} shows the results for
the claim categorization task (F1 macro over all classes; Tables
\ref{table_class_scores_1}--\ref{table_class_scores} in the Appendix
provide per-category results). Unsurprisingly, this fine-grained task
is more challenging for cross-lingual transfer. None of the
experimental configurations beats the monolingual baseline. As in
claim identification, translate-train outperforms multilingual
embeddings.

\paragraph{Error analysis.} Inspection of sentences shows that many misclassifications arise from
misleading local lexical material in the sentences. For example, “Die
SPD findet dies könnte die Integration unterstützen“ (``The Social
Democratic party believes this could support integration'') includes
the word 'integration' which is a strong cue for the claim category
'integration', which the model predicts. However, the correct category
is 'residency', as becomes clear from the broader context of the
article. Another example is: "Die sollen ja auch in der Gesellschaft
ankommen" (``They must arrive in society after all''), with misleading
cue 'society' indicating claim category 'society' and gold category
'integration'. A saliency analysis, as before, confirmed this pattern:
the “red herring” cues consistently receive the highest saliency
scores in the sentences. Notably, the error pattern persists in the
case of literal translations, but disppears when the translation
changes the wording (‘mit Sicherheit’ -- ``with security/certainty''
$\rightarrow$ ‘certainly’).

\section{Experiment 2: Cross-outlet cross-lingual transfer}
\label{sec:guardian-test-set}

%We now turn to our manually annotated English dataset,
%\textit{Guardian}. For comparability, also here we model claim
%identification at the sentence level.
%For the claim identification
%task, the dataset contains 1347 sentences in total, of which 112
%(8.3\%) containing at least a claim.

\begin{table}[tb!]
%\footnotesize
	\begin{center}
		\begin{tabular}{lrrll}
			\toprule
			Setup & Train & Test & Id & Cat \\
                        \cmidrule(r){1-3}
                        \cmidrule(l){4-5}
   			translate-train & en & en & \textbf{25.5} & 51.0 \\
                        \cmidrule(r){1-3}
                        \cmidrule(l){4-5}
			translate-test & de & de-en & 20.6 & \textbf{53.4} \\
                        \cmidrule(r){1-3}
                        \cmidrule(l){4-5}
			multilingual & de & en & 20.0 & 39.0 \\	
     \bottomrule
		\end{tabular}
		\caption{Guardian test set results for claim identification (Id, F1 of positive class) and claim categorization (Cat, macro F1)}\label{tabelle_guardian}
	\end{center}
\end{table}	

Results on the Guardian test set are shown in Table
\ref{tabelle_guardian}. For claim identification, the translate-train
approach outperforms the other approaches, confirming the trend seen
on the DebateNet data. For claim categorization, translate-test
outperforms translate-train and multilingual embeddings. Both of these
results are in line with our findings in Exp.\ 1.

For both tasks, we see a substantial decrease of performance on the
Guardian data (-30 points for claim identification, -15 points for
claim categorization). Since our previous experiment also used English
data, this performance drop cannot be due to cross-lingual differences, but
rather to differences between the two outlets, taz and the
Guardian. Indeed, we see that a British newspaper is likely to report
differently on German domestic affairs than a German newspaper, which
leads to differences in claim form and substance: They tend to focus
on the internationally most visible actors and report claims on a more
coarse-grained level. They also overreport the claim categories most
relevant for the British readership: claims migration control account
for 22\% of all claims in DebateNet but for 34\% in the Guardian. In
contrast, domestic (German) residency issues make up 14\% of the
DebateNet claims but only 2\% of the Guardian claims. See
Table~\ref{tabelle_classes} in the Appendix for a detailed breakdown
and example claims.

Thus, even if the Guardian claims might be structurally easier to
recognize, the cross-outlet differences in claim distribution make
transferring model representations from DebateNet to the Guardian
hard. The confusion matrix for claim identification in Table
\ref{tabelle_conf_guardian} shows a low-precision scenario, in
contrast to the high precision of the cross-lingual within-DebateNet
setup.

It is interesting to note that claim identification suffers much more
(-30 points) than claim categorization (-15 points), indicating that
the model of claim topics survives the transfer to another outlet
better than the model of what constitutes a claim.

\begin{table}[tb]
	\begin{center}
		\begin{tabular}{rrr}
			\toprule
			 & Target: yes & Target: no  \\
			\midrule
			Predicted: yes & 29 & 147  \\	
			Predicted: no & 83  & 1088 \\
			\bottomrule
		\end{tabular}
		\caption{Claim identification (Guardian): confusion matrix of the best model for English (translate-train)}\label{tabelle_conf_guardian}
	\end{center}
\end{table}

\section{Conclusion}

This paper explores different strategies for the cross-lingual
projection of political claims analysis from German into English and
French. Our experiments establish the potential of machine translation
for both claim identification and categorization, setting the stage
for further investigations on the factors affecting projection
performance and on the applicability of cross-lingual transfer for
similar analyses. Multilingual embeddings yielded worse results, in
line with previous analyses arguing that they attempt to solve a
harder (since more open-ended) task than Machine Translation
\cite{pires-etal-2019-multilingual,Barnes2019}. We find that the
language is not the only relevant dimension, though: in fact, the
differences in presentation between German and British articles on
German affairs go substantially beyond the language gap
\citep{VU2019101942}.

\section*{Acknowledgements}

This study was partially funded by Deutsche
Forschungsgemeinschaft (DFG) through MARDY (Modeling Argumentation
Dynamics) within SPP RATIO and by Bundesministerium für Bildung und
Forschung (BMBF) through E-DELIB (Powering up e-deliberation: towards
AI-supported moderation). We are grateful to Brandon Sorensen, who
annotated the Guardian test set.
%, and to the anonymous reviewers whose
%feedback helped improving the paper.

\section*{Limitations}

Our main experiment was limited to German, English, and French, three
typologically very similar languages. Generalization to more distant
languages is presumably harder, but was outside the scope of our
study.  Our Guardian test set is very small (albeit not significantly
smaller than out-of-domain gold sets often gathered for validation
purposes), and annotating it was challenging due to the need to apply
a codebook developed for the German debate to an English source. We
are currently working on improving the size and quality of our test
set.

While our experiments are reassuring as regards translation quality,
we cannot exclude that translation biases may have been introduced in
the data. We are also aware that DeepL is not the only option for
automatic translation; evaluating different translation methods,
however, falls outside the scope of this work.

\section*{Ethical Considerations}

 At the level of datasets and annotations, we employed an existing
 dataset (DebateNet2.0). Our own annotation contribution (the Guardian
 test set) was based on publicly available data; moreover, the
 annotation task was carried out following best practices. The Guardian test set is available upon request.

At the modeling level, we use previously defined models that are
publicly available; in this sense, our contribution does not raise new
ethical questions (e.g. in terms of misuse potential). To the
contrary, our focus is on understanding how these models transfer
across languages and what biases can potentially arise in this
transfer, as shown by our focus on error analysis.

% Entries for the entire Anthology, followed by custom entries
\bibliography{anthology,custom}
\bibliographystyle{acl_natbib}

\onecolumn
\appendix

\section{Appendix}
\label{sec:appendix}

\subsection{Datasets: quantitative details and comparison}

\begin{table*}[h]
\footnotesize
	\begin{center}
		\begin{tabularx}{\textwidth}{p{0.7cm}|p{3.8cm}|c|c|p{8.2cm}}
			%\toprule
			Class & Label & \%DN & \%G & Examples \\
			\midrule
			C1 & Controlling Migration & 22 & 34 & \textit{DN}: A fixed resettlement programme is needed, with binding annual admission quotas. \newline \textit{G}: Angela Merkel stressed the need for a fairer distribution of refugees across the EU  \\
   \midrule
			C2 & Residency & 14 & 2 & \textit{DN}: These urgent procedures shall be carried out in special reception facilities.  \newline \textit{G}: We have to find suitable accommodation for all of them. \\
   \midrule
			C3 & Integration & 9 & 3 & \textit{DN}: The CDU insists on an integration obligation for migrants. \newline \textit{G}: Michael Fuchs called on the government to set up language courses and to send job centre employees to assess newcomers \\
   \midrule
			C4 & Domestic Security & 3 & 8 & \textit{DN}: The head of the police union, RainerWendt, has called for a "ban mile around refugee shelters".  \newline \textit{G}: We should not hand over our streets to hollow rallying cries \\
   \midrule
			C5 & Foreign Policy & 16 & 11 & \textit{DN}: The current problems with the refugees must nevertheless be solved at European and international
level, she said. \newline \textit{G}: Tomas de Maizière said pressure should be applied to rejectionist nations such as Hungary, Slowakia and the Czech Republic. \\
   \midrule
			C6 & Economy + Labour Market & 3 & 7 & \textit{DN}: A condition for waiving such proof, however, must be that collective bargaining conditions or
a minimum wage apply in order to prevent dirty competition to the detriment of all employees. \newline \textit{G}: Folkerts-Landau said the influx of refugees has the potential not just to invigorate our economy but to protect prosperity for the future generations \\
   \midrule
			C7 & Society & 17 & 21 & \textit{DN}: And Reinhard Marx, chairman of the Catholic Bishops’ Conference, criticized the
strict separation between war refugees and economic refugees.  \newline \textit{G}: As chancellor, I come to the defense of Muslims, most of whom are upright, constitutionally loyal citizens \\
   \midrule
			C8 & Procedures & 15 & 14 & \textit{DN}: The federal government is planning a new law to speed up asylum procedures. \newline \textit{G}: Gerd Mueller called on Tuesday for the EU to appoint a European Refugees commissioner and said it had to treat the problem with more urgency \\
   \midrule
			%\bottomrule
		\end{tabularx}
		\caption{Claim categories: class, labels,  distribution (percentage of total claims), and example claim in DebateNet2.0 (DN) (manually translated into English) and Guardian test set (G).}\label{tabelle_classes}
	\end{center}
\end{table*}

\subsection{Per-category Results}

\begin{table}[h]
%\footnotesize
%test {1: 35, 2: 2, 3: 3, 4: 8, 5: 11, 6: 7, 7: 21, 8: 14, 9: 1}
	\begin{center}
		\begin{tabular}{lrrrc}
			\toprule
			Class & \#instances in test & Precision & Recall & F1 score \\
			\midrule
			C1 (Controlling Migration) & 35 & 0.67 & 0.83 & 0.74 \\
			C2 (Residency) & 2 & 0.66 & 0.74 & 0.70\\
			C3 (Integration) & 3 & 0.66 & 0.60 & 0.63\\
			C4 (Domestic Security) & 8 & 0.50 & 0.44 & 0.47\\
			C5 (Foreign policy) & 11 & 0.87 & 0.76 & 0.81\\
			C6 (Economy) & 7 & 0.88 & 0.50 & 0.64\\
			C7 (Society) & 21 & 0.70 & 0.67 & 0.69\\
			C8 (Procedures) & 14 & 0.75 & 0.70 & 0.72\\
			\midrule
			micro avg & & 0.71 & 0.71 & 0.71\\
			macro avg & & 0.71 & 0.66 & 0.67\\
            \bottomrule
		\end{tabular}
		\caption{Claim categorization: precision, recall and F1 values for the different classes, translate-train French} \label{table_class_scores_1}
	\end{center}
\end{table}	

\begin{table}[h]

	\begin{center}
		\begin{tabular}{lrrrc}
			\toprule
			Class & \#instances in test & Precision & Recall & F1 score \\
			\midrule
			C1 (Controlling Migration) & 35 & 0.66 & 0.74 & 0.70 \\
			C2 (Residency) & 2 & 0.68 & 0.70 & 0.69\\
			C3 (Integration) & 3 & 0.72 & 0.51 & 0.60\\
			C4 (Domestic Security) & 8 & 0.40 & 0.33 & 0.36\\
			C5 (Foreign policy) & 11 & 0.85 & 0.65 & 0.73\\
			C6 (Economy) & 7 & 0.80 & 0.57 & 0.67\\
			C7 (Society) & 21 & 0.77 & 0.56 & 0.65\\
			C8 (Procedures) & 14 & 0.76 & 0.58 & 0.66\\
			\midrule
			micro avg && 0.71 & 0.62 & 0.67\\
			macro avg && 0.70 & 0.58 & 0.63\\
            \bottomrule
		\end{tabular}
		\caption{Claim categorization: precision, recall and F1 values for the different classes, translate-train English} \label{table_class_scores_2}
	\end{center}
\end{table}	

\begin{table}[h]

	\begin{center}
		\begin{tabular}{lrrrc}
			\toprule
			Class & \#instances in test & Precision & Recall & F1 score \\
			\midrule
			C1 (Controlling Migration) & 35 & 0.76 & 0.71 & 0.73 \\
			C2 (Residency) & 2 & 0.76 & 0.69 & 0.72\\
			C3 (Integration) & 3 & 0.72 & 0.58 & 0.64\\
			C4 (Domestic Security) & 8 & 0.40 & 0.33 & 0.36\\
			C5 (Foreign policy) & 11 & 0.86 & 0.65 & 0.74\\
			C6 (Economy) & 7 & 0.83 & 0.36 & 0.50\\
			C7 (Society) & 21 & 0.86 & 0.56 & 0.68\\
			C8 (Procedures) & 14 & 0.73 & 0.61 & 0.66\\
			\midrule
			micro avg && 0.76 & 0.62 & 0.68\\
			macro avg && 0.74 & 0.56 & 0.63\\
            \bottomrule
		\end{tabular}
		\caption{Claim categorization: precision, recall and F1 values for the different classes, German baseline} \label{table_class_scores_3}
	\end{center}
\end{table}	

\begin{table}[h]

	\begin{center}
		\begin{tabular}{lrrrc}
			\toprule
			Class & \#instances in test & Precision & Recall & F1 score \\
			\midrule
			C1 (Controlling Migration) & 35 & 0.74 & 0.78 & 0.76 \\
			C2 (Residency) & 2 & 0.69 & 0.84 & 0.76\\
			C3 (Integration) & 3 & 0.72 & 0.62 & 0.67\\
			C4 (Domestic Security) & 8 & 0.48 & 0.61 & 0.54\\
			C5 (Foreign policy) & 11 & 0.81 & 0.81 & 0.81\\
			C6 (Economy) & 7 & 0.70 & 0.50 & 0.58\\
			C7 (Society) & 21 & 0.72 & 0.66 & 0.69\\
			C8 (Procedures) & 14 & 0.70 & 0.68 & 0.69\\
			\midrule
			micro avg && 0.72 & 0.73 & 0.72\\
			macro avg && 0.70 & 0.69 & 0.69\\
            \bottomrule
		\end{tabular}
		\caption{Claim categorization: precision, recall and F1 values for the different classes. Model: best cross-lingual model (translate-test)}\label{table_class_scores}
	\end{center}
\end{table}

\begin{table}

	\begin{center}
		\begin{tabular}{lrrc}
			\toprule
			Class & Precision & Recall & F1 score \\
			\midrule
			C1 (Controlling Migration) & 0.66 & 0.66 & 0.66 \\
			C2 (Residency) & 0.25 & 0.50 & 0.33\\
			C3 (Integration) & 0.50 & 0.67 & 0.57\\
			C4 (Domestic Security) & 1.00 & 0.25 & 0.40\\
			C5 (Foreign policy) & 0.45 & 0.82 & 0.58\\
			C6 (Economy) & 0.50 & 0.29 & 0.36\\
			C7 (Society) & 0.76 & 0.76 & 0.76\\
			C8 (Procedures) & 0.57 & 0.29 & 0.38\\
			\midrule
			micro avg & 0.61 & 0.58 & 0.60\\
			macro avg & 0.59 & 0.53 & 0.51\\
            \bottomrule
		\end{tabular}
		\caption{Claim categorization: precision, recall and F1 values for the different classes on Guardian dataset. Model: translate-test} \label{table_class_scores_guardian}
	\end{center}
\end{table}	

%Run 1: {'tp': 29, 'tn': 1088, 'fp': 147, 'fn': 83, 'f1': 0.201, 'acc': 0.82}
%Run 2: {'tp': 35, 'tn': 1050, 'fp': 185, 'fn': 77, 'f1': 0.210, 'acc': 0.81}

\end{document}